\documentclass{article}
\usepackage{ijcai15}

\usepackage{times}

\usepackage{graphicx}
\usepackage{amsmath}
\usepackage{amssymb}
\usepackage{algorithm, algorithmic}
\usepackage{multirow}

\def \bb {\mathbf b}
\def \ccc {\mathbf c}
\def \x {\mathbf x}

\def \z {\mathbf z}
\def \h {\mathbf h}

\def \0{{\mathbf 0}}
\def \1{{\mathbf 1}}

\def \D {\mathbf D}
\def \HH {\mathbf H}
\def \LL {\mathbf L}
\def \SSS {\mathbf S}
\def \W {\mathbf W}
\def \I {\mathbf I}
\def \J {\mathbf J}
\def \X {\mathbf X}

\def \MM {{\mathcal M}}
\def \GM {{\mathcal G}}
\def \tx {\tilde{\mathbf x}}

\def \st {\text{s.t.}}

\title{Learning Robust Features with Incremental Auto-Encoders}
\author{Yanan Li, Donghui Wang \\
Institute of Artificial Intelligence, Zhejiang University\\
\{dhwang, ynli\}@zju.edu.cn}
\begin{document}
\newcommand{\tabincell}[2]{\begin{tabular}{@{}#1@{}}#2\end{tabular}}
\maketitle

\begin{abstract}
  Automatically learning features, especially robust features, has attracted much attention in the machine learning community. In this paper, we propose a new method to learn non-linear robust features by taking advantage of  the data manifold structure. We first follow the commonly used trick of the trade, that is learning robust features with artificially corrupted data, which are training samples with manually injected noise. Following the idea of the auto-encoder, we first assume features should contain much information to well reconstruct the input from its corrupted copies. However, merely reconstructing clean input from its noisy copies could make data manifold in the feature space noisy. To address this problem, we propose a new method, called Incremental Auto-Encoders, to iteratively denoise the extracted features.  We assume the noisy manifold structure is caused by a diffusion process. Consequently, we reverse this specific diffusion process to further contract this noisy manifold, which results in an incremental optimization of model parameters . Furthermore, we show these learned non-linear features can be stacked into a hierarchy of features. Experimental results on real-world datasets demonstrate the proposed method can achieve better classification performances.
\end{abstract}

\section{Introduction}
Feature extraction, transforming the original input features to new feature space, has attracted  much attention in machine learning community, especially when data are represented by high dimensional feature vectors. Many linear (e.g. PCA, LDA, etc.) and non-linear feature learning methods (e.g. sparse coding, dictionary learning, etc.) have been proposed to address this problem during the past few years ~\cite{elad2010sparse,bengio2013representation}. Recent years, learning robust features is getting more and more attention from researchers in various areas, especially in the deep learning community ~\cite{wan2013regularization,farabet2013learning}.

In general, considering the types of training sets, robust feature learning methods can be roughly classified into two groups. Algorithms in one group learn features from natural noisy datasets. Whereas, methods in the other group are given clean training datasets.  In order to extract robust features, they learn with artificially corrupted data, which are training samples with manually injected noise.  For example, to learn robust features, handwritten digits dataset are manually injected with various noises, such as random binary background noise or image background noise \cite{larochelle2007empirical}.  In the deep learning literature, DAE (Denoising Auto-encoder) \cite{VincentLLBM10}, composed of an encoding and a decoding function,  is one of the best known building blocks for constructing a hierarchy of non-linear features. Features are made robust by reconstructing the clean input from its artificially corrupted copies via a decoding function. To make features invulnerable to different noises, AMC-SSDA combines multiple DAEs by a set of weights \cite{agostinelli2013adaptive}. Although the performances of these methods are prominent in many cases, their efficiency can also be improved since from the view of manifold learning, the high dimensional data are nearly lying on a low dimensional manifold. These methods have not taken fully considerations about the manifold structure.

To leverage the manifold structure,  some methods have been proposed. Typical linear feature learning methods are: LPP \cite{niyogi2004locality}, LLE \cite{roweis2000nonlinear} and Isomap \cite{tenenbaum2000global}. They learn linear features by preserving the local relationships within the data set and uncovering its essential manifold structure. There are also some other non-linear feature learning algorithms, such as SNE \cite{hinton2002stochastic}, t-SNE \cite{maaten2009learning}, etc. Commonly, these methods use various neighborhood graphs to characterize the manifold structure.  Differently, in the deep learning community, CAE (Contractive Auto-encoder) uses a contractive penalty term to force the learned features to capture the local direction of the non-linear manifold \cite{DBLP:conf/icml/RifaiVMGB11}. Compared with linear feature learning approaches, non-linear methods have proven to perform better in many cases. In addition, extracting robust features with the consideration of non-linear manifolds structures has also attracted much attention~\cite{DBLP:conf/nips/HeinM06,DBLP:conf/cvpr/WangT13}. However, these works learn the same dimensional features as the high dimensional input. In practice, not all features are relevant and important to the learning task, many of which are often redundant.

In light of these works, in this paper we propose a new method which can learn non-linear and robust features from manifold-embedded datasets. Similar to the above work, we first follow the well-known trick of the trade to learn with artificially corrupted data for extracting robust features. We assume extracted features should contain much information. Thus they can well reconstruct the clean input from its corresponding corrupted samples via a decoding function.  To get more reliable features, we then using a denoising method based on the following assumption: local structures of data in different features space should be consistent. From the view of manifold learning, artificially corrupting data makes the embedded manifold of input noisy. Merely minimizing the reconstruction error can not guarantee manifold in the new feature space being noiseless.  Thus manifolds are inconsistent with that of the input. To address this problem, we iteratively refine the learned features using a Laplacian-based method. We assume the noisy manifold is formed by a diffusion process on the Laplacian graph of data. We then reverse this diffusion process to denoise hidden features. Each step, representations are denoised towards the manifold. Step by step, the manifold structure of data becomes more and more refined. We further show that these non-linear features can be stacked to yield multiple levels of representations. Experimental results on several real world datasets illustrate that the proposed method can achieve better performance.

The remainder of the paper is organized as follows. We introduce the related work in Section 2. Then the proposed method  is presented in Section 3, followed by the optimization of the proposed method in Section 4. Following the experimental results in Section 5,  we conclude the paper in Section 6.

\section{Related Work}
In recent years, automatically feature learning has received increasing attention from machine learning community, especially in the deep learning community, due to its wide applications in practice. There are a rich body of work on feature learning in the literature. We provide a review to the most related methods in this section.

\textbf{Auto-encoder.} The auto-encoder is one of the most popular methods for learning informative non-linear features. It assumes these extracted features should contain as much information of input as possible and well reconstruct its input. To extract non-linear features, it exploits a direct parameterized function $f(\x)$, called encoder, to output hidden representations, defined as follows.
\begin{equation}
   \h = f(\x) = s_e(\W_1 \x + \bb_1)
   \label{eq:encoder}
\end{equation}
where $\W_1 \in \mathbb{R}^{K \times D}$ is the weight matrix and $\bb_1$ is the bias vector.  $\h$ is the $K$-dimensional feature vector.

In the meanwhile, another function $g(\h)$, called decoder, is defined to map from feature space back into the input space, producing a reconstruction $\hat{\x}$. It is parameterized as:
\begin{equation}
     \hat{\x}= g(\h) = s_d(\W_2 \h + \bb_2)
     \label{eq:decoder}
\end{equation}
where $\W_2 \in \mathbb{R}^{K \times D}$  and $\bb_2 \in \mathbb{R}^D$ are the weight matrix and bias vector respectively. $s_e$ and $s_d$ are the activation functions, whose typical choices are \emph{sigmoid, tanh, rectified linear.}

The set of parameters $\theta = \{\W_1, \W_2, \bb_1, \bb_2\}$ are learned simultaneously on the task of minimizing the reconstruction error over the whole training dataset $\X = \{\x_1, \x_2, ..., \x_n\} \in \mathbb{R}^{D \times n }$, which correspond to the following optimization function:
\begin{equation}
  \theta^{\star} = \arg\min_{\theta} \frac{1}{n}\sum_{i = 1}^n  l(\x_i, g(f(\x_i))),
 \label{eq:obj_ae}
\end{equation}
where $l$ is the reconstruction loss, whose typical choices are cross-entropy loss  and the squared error loss.

Traditionally, auto-encoder is used as a dimensionality reduction technique, which can learn equivalent or more useful features than what are obtained with simple linear PCA. Recently, a more successful use of auto-encoder is to learn over-complete features, yielding more rich hidden representations. However, this renders the problem that the basic auto-encoder can learn an identity mapping with perfectly reconstructing its input and without extracting more meaningful features. To tackle this problem, various methods with different criteria have been proposed, such as sparse auto-encoder \cite{ngiam2011optimization}, RBM \cite{hinton2006fast} and so on. Among all the various constraints, robustness of features is most favored.

\textbf{Denoising Auto-encoder.} One popular method to impose the robustness constraint is denoising auto-encoder (DAE). Except for remaining much information of input, it assumes good hidden features should well reconstruct its clean input from the corrupted copies, which avoids the uninteresting solutions of auto-encoder. From the geometric structure of input, which assumes high dimensional data are concentrated on a low dimensional manifold $\MM$, DAE maps far away corrupted data to small regions close to the intrinsic data manifold. Formally, it is trained by the following function:
\begin{equation}
     \theta^{\star} = \arg\min_{\theta} \frac{1}{nm}\sum_{i = 1}^n \sum_{j = 1}^m l(\x_i, g(f(\tx_{i_j}))),
     \label{eq:obj_dae}
\end{equation}
where each sample $\x_i$ is reconstructed from its $m$ corrupted copies $\tx_{i_j} = \rho(\x_i)$. Typical choices for the corrupting function $\rho$ are \emph{additive isotropic Gaussian noise, salt and pepper noise and masking noise}.  Comparing to the traditional auto-encoder, these learned features are qualitatively better in classification performance. Exploiting DAE as a building block, several other methods have been proposed, such as AMC-SDAE \cite{agostinelli2013adaptive}, spDAE \cite{cho2013simple}, mDAE \cite{chen2014marginalized} and so on.

However, DAE still subjects to some drawbacks. Based on the manifold hypothesis, hidden representations correspond to an intrinsic coordinate system on the manifold structure. Variations in the input should be reflected in the learned representation. Whereas, since DAE makes the whole mapping robust instead of $\h$, this assumption is not guaranteed. In addition, just mapping back corrupted samples to a nearby region makes the intrinsic manifold structure divergent. It fails to maintain the local structure when multiple manifolds exists in training data, which is often the case.

\textbf{Contractive Auto-encoder.} Another method to learn robust features is contractive auto-encoder (CAE). From a different perspective, it assumes features should be contractive along the orthogonal direction to the manifold. Its goal is achieved by adding a contractive penalty term directly on the hidden features to the basis auto-encoder. Hidden features are made insensitive to small changes of input by the Frobenius norm of the encoder's Jocabian. It is trained by minimizing the following objective function:
\begin{equation}
    \theta^{\star} = \arg\min_{\theta} \sum_{i=1}^n  l(\x_i, g(f(\x_i))) + \lambda ||\J(\x_i)||^2_F
  \label{eq:obj_cae}
\end{equation}
where $\J \in \mathbb{R}^{K \times D}$ is the encoder's Jacobian matrix and $\lambda$ is the trade-off parameter.

Comparing with DAE, CAE captures the local changes of the data manifold in the hidden representation. However, the contractive penalty term merely encourages robustness to infinitesimal changes of input. Thus, when data is corrupted by a large noise, it could fail. This problem is further considered by \cite{rifai2011higher}, which penalizes all higher order derivatives.

\section{InAE: Incremental Auto-Encoders}
\subsection{Problem modeling}
From the manifold hypothesis, hidden representations correspond to an intrinsic coordinate system on the embedded manifold $\MM$.  Variations along the manifold in the input space should be well captured or reflected in the learned representations. However, merely reconstructing clean input from itself or its noisy copies could make manifold $\MM$ in the feature space noisy. As a result, intrinsic manifolds between original space and the hidden feature space are not consistent.
To converge the noisy manifold, DAE uses a denoising criterion while CAE proposes a contractive penalty. Differing from them, we refine the manifold structure by reversing a diffusion process, which results in an incremental optimization of the model parameter.

\subsection{Reverse the diffusion process to contract the noisy manifold}
The data manifold in the extracted feature space is noisy, which is obtained by learning with artificially corrupted data, i.e. training samples with manually injected noise. We assume the divergent manifold is caused by  a diffusion process from the intrinsic manifold $\MM$.  Consequently, we propose to reverse the specific diffusion process to refine the manifold structure .

Formally, given the noisy hidden features $\HH = \{\h_1, \h_2, ..., \h_{nm}\}$ , we reverse the diffusion process iteratively by the following equation:
\begin{equation}
  \partial_t \HH = -\gamma \LL \HH,
  \label{eq:MD}
\end{equation}
where $\gamma$ is the diffusion constant and $t$ indicates the $t$-th iteration. $\LL = (\D - \SSS)\in \mathbb{R}^{N \times N}$ is the Laplacian matrix of $\GM$, where $\SSS$ is the similarity matrix of $\X$ and $\D_{ii} = \sum_{j=1}^N \SSS_{ij}$. Along with the increase in the number of iteration, $\HH$ is inching closer to the data manifold $\MM$.

Using an implicit Euler-scheme, $\HH$ is updated by the following function,
\begin{equation}
  \HH^{t+1} = (\I + \delta t \gamma \LL )^{-1} \HH^t,
  \label{eq:diffusion}
\end{equation}
where $\delta t$ is the time-step and can be chosen arbitrarily.  We assume step by step, $\h$  goes closer to $\MM$.

\subsection{Adaptively construct the neighboring graph }
\label{section:graph}
During the process of reversing the diffusion process, its generator, i.e. the graph Laplacian of the neighborhood graph is a key factor.  Similarly, it does matter how the neighborhood graph is constructed. A good neighborhood graph can potentially preserve the locality of data manifold. Here we use two alternative strategies to construct the neighboring graph $\GM$.

The first one is the popular method $k$-nn, which chooses $k$ nearest neighbors for data $\x_i$.  $k$-nn performs pretty well in most cases. However, several problems still arises with $k$-nn, especially when data are concentrated on a non-linear manifold.  (1) $k$-nn assumes data are distributed over a Gaussian distribution. This is often violated by real world data which are concentrated over a complex non-linear manifold. Most Euclidean nearest neighbors are chosen from different data manifolds. (2) When clusters have unbalanced number of training samples, it is not proper to set the same value of $k$ for different clusters.  For each data, it would be better to choose its  nearest neighbors automatically according to the intrinsic manifold.

Thus, an alternative method is proposed.  First, we utilize $k$-nn to choose relatively large number of neighbors for each data point $\x$. These neighbors, denoted as $\X_G(\x)$, come from not only the same manifold as $\x$, but also different manifolds of other classes. Then we explore a sparse subspace learning method \cite{DBLP:journals/corr/abs-1203-1005}  to further select these neighbors. The sparse subspace clustering method has turned out to be very effective for discovering data manifold in high dimensional space. It assumes each data $\x$ is a linear combination of its neighbors within the same cluster. By optimizing a $\ell_1$ minimization problem, samples with non-zero coefficients are adaptively selected as neighbors.  Thus,  the nearest neighbors are finally selected by optimizing the following function:
\begin{equation}
    \arg \min_{\ccc} ||\x- \X_G(\x) \ccc|| + \lambda ||\ccc||_1,
    \label{eq:neighbor}
\end{equation}
where $\ccc \in \mathbb{R}^k$ is the corresponding sparse coefficient. Points with the non-zeros coefficients, are treated as the neighbors of $\x$, denoted as $\X_L(\x)$. Combining these two steps can not only select effective neighbors from the same manifold, but remove the neighbors lying in the different manifolds, as the experimental results demonstrate.

Thus, the similarity matrix $\SSS$ can be constructed as follows:
\begin{equation}
    \SSS_{ij} = \left\{
      \begin{array}{ll}
        d(\x_i, \x_j)& \text{if} \quad \x_j \in \X_L(\x_i)\\
       0 & \text{else}
      \end{array}
      \right.
      \label{eq:simiS}
\end{equation}
where $d(\x_i, \x_j)$ measures the similarity between $\x_i$ and $\x_j$, which can be chosen as Gaussian kernel $\exp^{-\frac{||\x_i - \x_j||^2}{2 \sigma^2}}$ or cosine distance $\frac{\x_i^T \x_j}{||\x_i||||\x_j||}$ \cite{yan2007graph}.

\subsection{ Impose insensitivity to input noise}
To further learn robust features, we follow the same idea in DAE, i.e. features should  well reconstruct clean input from its corrupted copies . In each step, we assume the hidden feature $\h$  should: (1) contain much information of the input and well reconstruct $\x$ from its corrupted versions $\tx$. (2) approach the intrinsic manifold gradually, i.e. manifold of $\HH$ is being gradually denoised.

Thus, we formulate the proposed method, incremental auto-encoder (InAE) is obtained by combing Eq.\ref{eq:obj_dae} with Eq.\ref{eq:diffusion}. In each step, features are learned by optimizing the following objective function.
\begin{equation}
\begin{split}
   \theta^{\star}_{t+1} = \arg \min _{\theta_{t+1}}  \frac{1}{nm}\sum_{i = 1}^n \sum_{j = 1}^m l(\x_i , g(f(\tilde{\x}_{i_j}))),\\
    \st \  \HH^{t+1} = (\I + \delta t \gamma \LL )^{-1} \HH^t,
\end{split}
   \label{eq:obj1}
\end{equation}
where $\tilde{\x}_{i_j}$ is the $j$-th noisy copy of $\x_i$ and $\HH^t = f({\W_1^t} \tilde{\X} + \bb_1^t)$.

From the objective function Eq.\ref{eq:obj1}, we see: (1) the proposed method explicitly constrains the extracted features, which are prompted to capture the variations of the input; (2) Differing from CAE-like methods, the noise magnitude is not confined to infinitesimal. Thus robustness of features is guaranteed.

\section{Optimization of the objective function}
To train this model, we rewrite Eq.\ref{eq:obj1} as a general regularized function. Following the idea in \cite{scherzer2000relations}, Eq.\ref{eq:diffusion} is equivalent to the solution of the minimization of the following regularization problem:
\begin{equation}
   \Phi(\HH^{t+1}) = ||\HH^{t+1} - \HH^t||^2_F + (\delta t) tr(\HH^{t+1} \LL {\HH^{t+1}}^T),
    \label{eq:diffusion2}
\end{equation}
$tr(\cdot)$ computes the trace value.

Thus, the objective function in each step becomes:
\begin{equation}
   \theta^{\star}_{t+1}= \arg \min _{\theta_{t+1}}\frac{1}{nm}\sum_{i = 1}^n \sum_{j = 1}^m l(\x_i , g(f(\tilde{\x}_{i_j})))+ \alpha\Phi(\HH^{t+1}),
   \label{eq:obj_final}
\end{equation}
where $\alpha$ is the trade-off parameter between the reconstruction error and the process of reversing diffusion process. $\theta_{t+1} = \{\W_1^{t+1}, \W_2^{t+1}, \bb_1^{t+1}, \bb_2^{t+1}\}$ contains all the parameters in $(t+1)$-th iteration.

By analyzing the two penalty terms in Eq.\ref{eq:diffusion2}, we see (1) two consecutive updates of $\HH$ are forced to change smoothly. In other words, $\h$ comes gradually closer to the manifold $\MM$, which evades the oscillation phenomenon when optimizing the objective function. (2) close-by points in the original space is rendered to be close in the new feature space. Since  $tr(\HH^{t+1} \LL {\HH^{t+1}}^T) = \sum_{i,j}^N ||\h^{t+1}_i - \h^{t+1}_j||^2 \SSS_{ij}$,  if $\x_i$ and $\x_j$ are close, i.e. $\SSS_{ij}$ is large, $\h_i$ and $\h_j$ should be close as well. Specifically, the local structure in the data can be maintained.

Different from the traditional auto-encoder, the proposed method is trained by an incremental optimization procedure, resulting in a series of parameter updates:
\begin{equation}
  \HH^0 \rightarrow \theta_1\rightarrow \HH^1 \rightarrow  \theta_2 \ldots \HH^{T-1} \rightarrow \theta_T,
\end{equation}
where $T$ denotes the number of update. Each parameter $\theta_{t+1}$ is better than the last update $\theta_{t}$.


To obtain each parameter $\theta$, Eq.\ref{eq:obj_final} is optimized by stochastic gradient descent. Here, we just give a simple description of the first derivative of the last penalty. For clarity, we omit the subscript $t+1$ and denote the whole number of $\HH$ as $N$.  First, we compute the derivative w.r.t. each element $\W_1^{ij}$.
\begin{equation}
  \begin{split}
    &\frac{\partial tr(\HH \LL \HH^T)}{\partial \W_1^{ij}} = \frac{\partial \sum_{m,n}^N||\h_m - \h_n||^2\SSS_{mn}}{\partial \W_1^{ij}} \\
    & = \frac{\partial \sum_{m,n}^N \SSS_{mn}(\h_m^T\h_m + \h_n^T\h_n - 2\h_m^T\h_n)}{\partial \W_1^{ij}}\\
    & = \frac{2\sum_m^N \D_{mm} \partial \h_m^T\h_m}{\partial \W_1^{ij}} - \frac{2\sum_{m,n}^N \SSS_{mn}\partial \h_m^T\h_n}{\partial \W_1^{ij}}
  \end{split}
  \label{eq:derivative}
\end{equation}

As $\h_m$ is a non-linear function of $\z_m = \W_1 \x_m + \bb_1$, using the chain rule, Eq.(\ref{eq:derivative}) is written as:
\begin{equation}
4\sum_m^N(\D_{mm} \h_{m_j} - \sum_{n}^N \W_{mn}\h_{n_j})\frac{\partial \h_{m_j}}{\partial \z_{m_j}} \x_{m_i},
\end{equation}
where $\h_{m_j}$ is the $j$-th element in $\h_m$.

Therefore, we get the derivative w.r.t. $\W_1$ as follows:
\begin{equation}
\begin{split}
  &\frac{\partial tr(\HH \LL \HH^T)}{\partial \W_1}\\
  & = 4 \sum_{m}^N \{ \x_m[(\D_{mm} \h_m  - \sum_{n}^N\W_{mn}\h_n)^T \circ s'(\z_m)]^T\}
\end{split}
\end{equation}
where $\circ$ is the Hadamard product. In this paper, we use the sigmoid function in the encoder, where $s'(x) = s(x)(1-s(x))$.

In summary, the whole training algorithm is described in Algorithm. \ref{algorithm:1}
\begin{algorithm}[htb]
  \caption{Training the Incremental Auto-Encoder}
  \begin{algorithmic}[1]
    \REQUIRE Training data $\X$,  parameter $k$ and $\sigma$ in constructing $\W$, number of iterations $T$;
    \ENSURE Model parameters $\theta = \{\W_1$, $\W_2$, $\bb_1$ , $\bb_2\}$;
    \STATE Generate the noisy training dataset $\tilde{\X}$ from $\X$;
    \STATE Construct an adaptive neighborhood graph $\GM$ on training data;
    \STATE Compute similarity matrix $\SSS$ on $\GM$ by Eq.\ref{eq:simiS};
    \STATE Compute the Laplacian matrix $\LL = \D - \SSS$;
    \STATE Initialize $\HH^0$;
    \FOR{each iteration $t$}
       \STATE Update $\theta_t$ by stochastic gradient descent;
       \STATE Obtain $\HH^t = f({\W_1^t \tilde{\X} + \bb_1^t})$;
    \ENDFOR
    \STATE $\theta = \theta_T$;
  \end{algorithmic}
  \label{algorithm:1}
\end{algorithm}

\begin{figure*}[t]
\begin{center}
   \includegraphics[width=\linewidth]{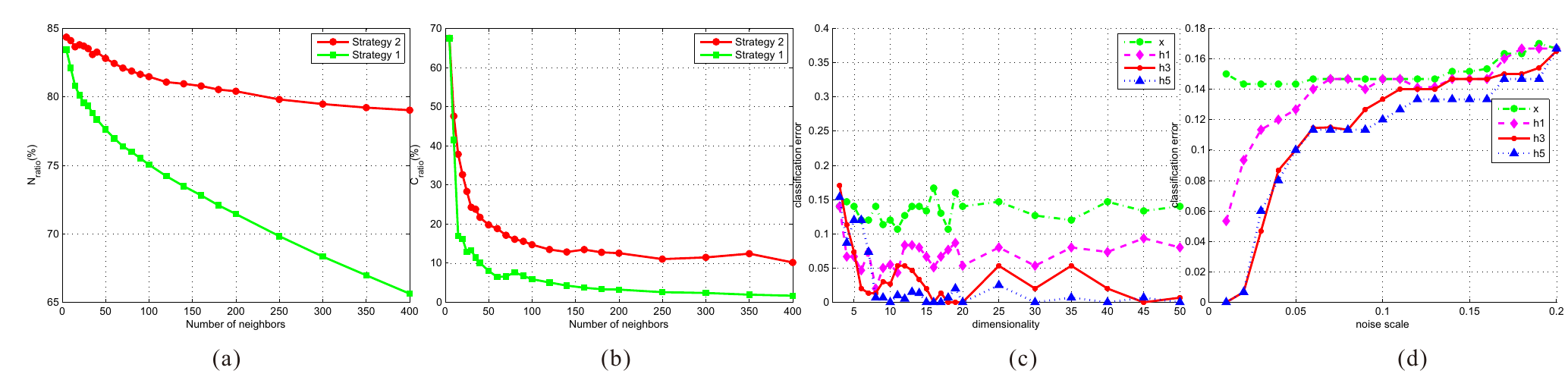}
\end{center}
   \caption{(a) and (b) are two indicators of locality preserving, $N_{ratio}$ and $C_{ratio}$. They indicates the selection of neighbors from the same manifold. Bigger is better. (c) and (d) give the classification error of different level representations with different dimensions and noise scale. $h1$ denotes denoised representations in the first time-step. We observe that as the increase of iteration number, hidden representations become more discriminative.}
\label{fig:toydata2}
\end{figure*}
\begin{figure*}[t]
\begin{center}
   \includegraphics[width=\linewidth]{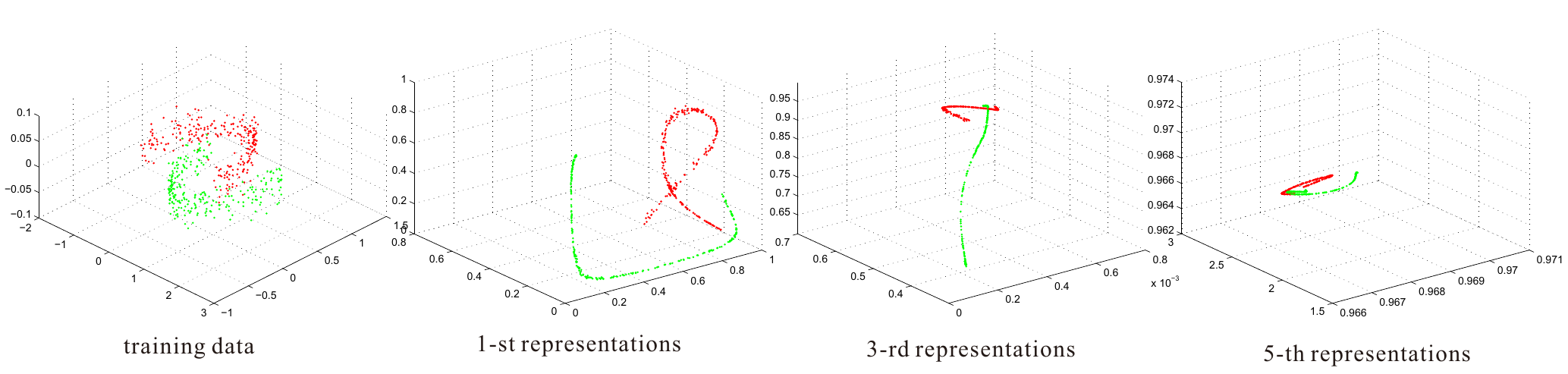}
\end{center}
   \caption{The original noisy 2 moon-like datasets and its denoised representations in 1-st, 3-rd and 5th time-step are displayed respectively. It is noticeable that the noisy manifold structure is denoised efficiently even with 1 update.}
\label{fig:toydata}
\end{figure*}
\subsection{Multiple levels of representation}
Similar to the auto-encoder, we treat the proposed method as a building block of forming the deep architecture.  Stacking several layers to initialize a deep network works in much the same way as stacking auto-encoders. We stack multiple layers by feeding the output $\h$ of $l^{th}$ layer as input into the $(l+1)^{th}$ layer. Once one layer is trained, the encoding function $f$ is used to generate uncorrupted input for the next layer.  Once a deep architecture has thus been built, its highest level output representation can be used as input to a stand-alone supervised learning algorithm. Experimental results show that the high level representations achieve better performance.

\section{Experimental Results}
In this section,
we separate the experiments into
model validation on synthetic data
and performance comparison on benchmark datasets,
showing the prominent locality preserving and discriminative performance
of our proposed method.

\subsection{Model validation on synthetic data}

{\bf Dataset.}
To validate our proposed method,
we generate a moon-like dataset consisting of 2 clusters,
each of which is generated from a 2-D function
and embedded into a 9-D space with an isotropic Gaussian noise $\epsilon\sim N(\0,\sigma\I)$.

{\bf Evaluation metric.}
We evaluate our method
on the locality preserving ability in two indicators $N_{ratio}$ and $C_{ratio}$,
and on discriminative power in classification error.

Since the number of neighbours is adaptive,
we choose the first $k$ neighbours corresponding to the number in $k$-nn algorithm.
For each data $\x_i$,
we introduce $N_{ratio}$ as the ratio of the number of selected neighbours on same manifold of $\x_i$ to $k$
and $C_{ratio}$ indicates the percentage of sum of the coefficients of selected neighbours on same manifold.
And large values mean good locality preserving ability.
The error is compared when $k$ is $30$ and iterations number $t$ is 5.

{\bf Experimental results.}
In order to show the performance of the proposed InAE,
we give simple illustrations in Fig.\ref{fig:toydata2}.

Fig.\ref{fig:toydata2} (a) and (b) illustrate results on indicators $N_{ratio}$ and $C_{ratio}$ of the two strategy for constructing neighboring graph, as described in Section \ref{section:graph}.
We see in (a) when $k$ is small,
$k$-nn and our method perform almost equally well,
since the data is relatively large.
The difference in $N_{ratio}$ grows larger as the number of neighbours is larger.
Similar result of $C_{ratio}$ in (b) further convinces the locality preserving ability of our method.

Fig.\ref{fig:toydata2} (c) and (d) show the errors
respect to dimensionality and noise scale on several numbers of iterations $t$.
We notice from (c) that
higher level representation is more discriminative,
and our model is more suitable when the dimension is higher.
And (d) shows when Gaussian noise scale $\sigma$ is small,
higher level representations are already able to obtain high classification performance.
However,
large $\sigma$ destroys data severely,
causing high classification error.

\begin{table*}[!htbp]
\caption{Test error rate(\%) on the MNIST variant datasets.  Other methods are trained with 1 hidden layer (left column) and 2 hidden layers (hidden layers). The left column indicates the result with 1 hidden layer, while the right column is that with 2 hidden layers.}
\centering
  \begin{tabular}{|c|c|c|c|c|c|c|c|c|c|c|c|}
  \hline
   &raw + SVM& \multicolumn{2}{|c|}{AE+ SVM} & \multicolumn{2}{|c|}{RBM+ SVM} & \multicolumn{2}{|c|}{DAE+ SVM} &\multicolumn{2}{|c|}{CAE+SVM}& \multicolumn{2}{|c|}{InAE+ SVM}\\
   \hline
   \hline
   \emph{mnist-rot} & 15.3 & 14.78 & 12.0 & 14.75 & 11.78& 14.02& 11.87& 11.50 & 10.02& 10.12& \textbf{9.89} \\
   \hline
    \emph{ mnist-back-rand} & 29.78 & 17.04& 13.45& 11.75& \textbf{9.50}& 14.53& 12.17&13.05& 11.03& 12.73& 10.75\\
    \hline
    \emph{mnist-back-image}& 29.27& 28.78& 27.01& 21.46& 20.3& 20.4& 19.9& 18.75& 17.82&18.01& \textbf{17.73} \\
    \hline
     \emph{mnist-rot-back-image} & 67.30& 55.47& 54.3& 53.85& 52.15& 51.31& \textbf{50.09}& 55.1& 53.95& 54.6& 53.9 \\
     \hline
     \emph{rectangles}& 2.19& 2.59& 2.46& 2.57& 2.35& 2.41& 2.18&2.02& 1.85&  1.39& \textbf{1.23} \\
     \hline
     \emph{rectangles-image} & 24.7& 26.7& 24.5& 24.9& 23.3& 23.5& 21.7&22.03& 21.05&  21.2& \textbf{20.9} \\
     \hline
     \emph{convex} & 29.8& 31.2& 29.8& 29.1& 28.5& 30.5& 29.3& 29.8& 28.9& 29.3& \textbf{28.4}\\
     \hline
  \end{tabular}
  \label{tab:bb}
\end{table*}

Fig.\ref{fig:toydata} gives an intuitive denoised results $\HH$,
with $t$ increases,
$\HH$ become cleaner and preserve their original manifold at the same time.
From Fig.\ref{fig:toydata2} and Fig.\ref{fig:toydata}, we see just 2 or 3 iterations is sufficient to greatly improve the classification accuracy of hidden features.
The discrimination power is strengthened at the cost of small computational complexity.

\subsection{Performance on benchmark datasets}

{\bf Dataset.}
After the validation of our model on synthetic data,
we compare our method with state-of-the-art algorithms on several popular benchmark datasets,
i.e. the handwritten digits MNIST dataset and its variants, and CIFAR-10 dataset.
The variants of MNIST are generated by imposing various challenging factors~\cite{larochelle2007empirical}.
The CIFAR-10 dataset consists of 6000 $32\times32$ color images in 10 classes.

{\bf Evaluation metric.}
Besides the traditional classification error metric on synthetic data,
we employ another measure $eig(\SSS^{-1}_w\SSS_b)$,
which denotes the maximum eigenvalue of $\SSS^{-1}_w\SSS_b$,
where $\SSS_w$ and $\SSS_b$ are the with-in-class and between-class variance of $\h$, respectively.
This measure is inspired from Fisher LDA,
which assumes that
a discriminative feature should make data in different classes far away,
while data in same class close to each other.
Large value indirectly indicates the discriminative power of the hidden representations,
and better classification performance is expected.

\begin{table}[h]
\caption{Performance comparison of all methods on MNIST and CIFAR-10 datasets.
$eig(\SSS_w^{-1}\SSS_b)$ here indicates its maximum eigenvalue (in $10^2$).
In this table,
the larger maximum eigenvalue means better classification performance.}
\begin{center}
\begin{tabular}{c|c|c|c|c|}
\cline{2-5}
& \multicolumn{2}{|c|}{MNIST} & \multicolumn{2}{c|}{CIFAR-10} \\
\cline{2-5}
& error & $eig(\SSS_w^{-1}\SSS_b)$ & error & $eig(\SSS_w^{-1}\SSS_b)$ \\
\hline
\cline{1-5}
\multicolumn{1}{|c|}{RBM} & 1.64 & 5.12 & - & - \\
\cline{1-5}
\multicolumn{1}{|c|}{AE} & 1.85 & 4.02 & 57.3 & 0.27 \\
\cline{1-5}
\multicolumn{1}{|c|}{DAE} & 1.43 & 6.32 & 51.8 & 0.30 \\
\cline{1-5}
\multicolumn{1}{|c|}{CAE} & 1.25 & 7.95 & 48.85 & 0.43 \\
\cline{1-5}
\multicolumn{1}{|c|}{InAE} & \textbf{1.12} & 8.68 & \textbf{47.75} & 0.59 \\
\cline{1-5}
\end{tabular}
\end{center}
\label{tab:aa}
\end{table}

{\bf Experimental results.}
We compare our method against the following algorithms on feature extraction:
AE (traditional auto-encoder),
DAE-b (denosing auto-encoders with masking-out noise),
CAE (contractive auto-encoders)
and RBM (restricted Boltzmann machine).
We use a linear SVM on the raw image pixels as baseline.
For all methods but RBM,
we use untied weights (i.e. $\W_1\neq\W_2$) in each layer
and train them using Stochastic Gradient Descent.
RBM is trained using Contrastive Divergence,
of which the hyper-parameter are chosen by a grid search on a validation set.

We represent the classification error and $eig(\SSS^{-1}_w\SSS_b)$
on MNIST and CIFAR-10 datasets in Tab.\ref{tab:aa},
and error rate on MNIST variant datasets in Tab.\ref{tab:bb}.
Our proposed method achieves best performance in almost all datasets,
and Tab.\ref{tab:aa} proves that the metric $eig(\SSS^{-1}_w\SSS_b)$ is positively correlated with classification performance.
We draw a conclusion that distance between samples provides features valuable information for subsequent tasks.
Since the manifold structure of data is important for good performance,
you may try to well utilize it before we have a chance.
Moreover,
Tab.\ref{tab:bb} demonstrates that stacking multiple layers significantly improves performance.

\section{Conclusions}
In this paper, we proposed a novel robust feature learning method by utilizing the Laplacian structure of training data. To learn robust features, it follows the well known trick in machine learning and learns with artificially corrupted data, which are training samples with manually injected noise. First, we assume features should contain much information and well reconstruct the clean input. However, since the data manifold is injected by some noise, this structure can not be consistent in the new feature space, if features are learned merely based on minimization of the reconstruction error. To address this problem, we model the noisy manifold is the result of a diffusion process on the Laplacian graph of training data. Then we reverse this specific diffusion process to denoise the manifold. Each time the diffusion process is reversed, the manifold is refined. This results in an incremental optimization of model parameters.  In addition, a new strategy of constructing the neighboring graph of data is introduced. We find that the Incremental Auto-Encoder is capable of contracting the noisy manifold in the feature space.  Experimental results on real-world datasets suggest that the Incremental Auto-Encoder performs better than other comparing methods.

\bibliographystyle{named}
\bibliography{ijcai15}

\begin{thebibliography}{}

\bibitem[\protect\citeauthoryear{Agostinelli \bgroup \em et al.\egroup
  }{2013}]{agostinelli2013adaptive}
Forest Agostinelli, Michael~R Anderson, and Honglak Lee.
\newblock Adaptive multi-column deep neural networks with application to robust
  image denoising.
\newblock In {\em Advances in Neural Information Processing Systems}, pages
  1493--1501, 2013.

\bibitem[\protect\citeauthoryear{Bengio \bgroup \em et al.\egroup
  }{2013}]{bengio2013representation}
Yoshua Bengio, Aaron Courville, and Pascal Vincent.
\newblock Representation learning: A review and new perspectives.
\newblock {\em Pattern Analysis and Machine Intelligence, IEEE Transactions
  on}, 35(8):1798--1828, 2013.

\bibitem[\protect\citeauthoryear{Chen \bgroup \em et al.\egroup
  }{2014}]{chen2014marginalized}
Minmin Chen, Kilian~Q Weinberger, Fei Sha, and Yoshua Bengio.
\newblock Marginalized denoising auto-encoders for nonlinear representations.
\newblock In {\em Proceedings of the 31st International Conference on Machine
  Learning (ICML-14)}, pages 1476--1484, 2014.

\bibitem[\protect\citeauthoryear{Cho}{2013}]{cho2013simple}
KyungHyun Cho.
\newblock Simple sparsification improves sparse denoising autoencoders in
  denoising highly noisy images.
\newblock ICML, 2013.

\bibitem[\protect\citeauthoryear{Elad}{2010}]{elad2010sparse}
Michael Elad.
\newblock {\em Sparse and redundant representations: from theory to
  applications in signal and image processing}.
\newblock Springer Science \& Business Media, 2010.

\bibitem[\protect\citeauthoryear{Elhamifar and
  Vidal}{2012}]{DBLP:journals/corr/abs-1203-1005}
Ehsan Elhamifar and Ren{\'{e}} Vidal.
\newblock Sparse subspace clustering: Algorithm, theory, and applications.
\newblock {\em CoRR}, abs/1203.1005, 2012.

\bibitem[\protect\citeauthoryear{Farabet \bgroup \em et al.\egroup
  }{2013}]{farabet2013learning}
Clement Farabet, Camille Couprie, Laurent Najman, and Yann LeCun.
\newblock Learning hierarchical features for scene labeling.
\newblock {\em Pattern Analysis and Machine Intelligence, IEEE Transactions
  on}, 35(8):1915--1929, 2013.

\bibitem[\protect\citeauthoryear{Hein and Maier}{2006}]{DBLP:conf/nips/HeinM06}
Matthias Hein and Markus Maier.
\newblock Manifold denoising.
\newblock In {\em Advances in Neural Information Processing Systems 19,
  Proceedings of the Twentieth Annual Conference on Neural Information
  Processing Systems, Vancouver, British Columbia, Canada, December 4-7, 2006},
  pages 561--568, 2006.

\bibitem[\protect\citeauthoryear{Hinton and
  Roweis}{2002}]{hinton2002stochastic}
Geoffrey~E Hinton and Sam~T Roweis.
\newblock Stochastic neighbor embedding.
\newblock In {\em Advances in neural information processing systems}, pages
  833--840, 2002.

\bibitem[\protect\citeauthoryear{Hinton \bgroup \em et al.\egroup
  }{2006}]{hinton2006fast}
Geoffrey Hinton, Simon Osindero, and Yee-Whye Teh.
\newblock A fast learning algorithm for deep belief nets.
\newblock {\em Neural computation}, 18(7):1527--1554, 2006.

\bibitem[\protect\citeauthoryear{Larochelle \bgroup \em et al.\egroup
  }{2007}]{larochelle2007empirical}
Hugo Larochelle, Dumitru Erhan, Aaron~C. Courville, James Bergstra, and Yoshua
  Bengio.
\newblock An empirical evaluation of deep architectures on problems with many
  factors of variation.
\newblock In {\em Machine Learning, Proceedings of the Twenty-Fourth
  International Conference {(ICML} 2007), Corvallis, Oregon, USA, June 20-24,
  2007}, pages 473--480, 2007.

\bibitem[\protect\citeauthoryear{Maaten}{2009}]{maaten2009learning}
Laurens Maaten.
\newblock Learning a parametric embedding by preserving local structure.
\newblock In {\em International Conference on Artificial Intelligence and
  Statistics}, pages 384--391, 2009.

\bibitem[\protect\citeauthoryear{Ngiam \bgroup \em et al.\egroup
  }{2011}]{ngiam2011optimization}
Jiquan Ngiam, Adam Coates, Ahbik Lahiri, Bobby Prochnow, Quoc~V Le, and
  Andrew~Y Ng.
\newblock On optimization methods for deep learning.
\newblock In {\em Proceedings of the 28th International Conference on Machine
  Learning (ICML-11)}, pages 265--272, 2011.

\bibitem[\protect\citeauthoryear{Niyogi}{2004}]{niyogi2004locality}
X~Niyogi.
\newblock Locality preserving projections.
\newblock In {\em Neural information processing systems}, volume~16, page 153,
  2004.

\bibitem[\protect\citeauthoryear{Rifai \bgroup \em et al.\egroup
  }{2011a}]{rifai2011higher}
Salah Rifai, Gr{\'e}goire Mesnil, Pascal Vincent, Xavier Muller, Yoshua Bengio,
  Yann Dauphin, and Xavier Glorot.
\newblock Higher order contractive auto-encoder.
\newblock In {\em Machine Learning and Knowledge Discovery in Databases}, pages
  645--660. Springer, 2011.

\bibitem[\protect\citeauthoryear{Rifai \bgroup \em et al.\egroup
  }{2011b}]{DBLP:conf/icml/RifaiVMGB11}
Salah Rifai, Pascal Vincent, Xavier Muller, Xavier Glorot, and Yoshua Bengio.
\newblock Contractive auto-encoders: Explicit invariance during feature
  extraction.
\newblock In {\em Proceedings of the 28th International Conference on Machine
  Learning, {ICML} 2011, Bellevue, Washington, USA, June 28 - July 2, 2011},
  pages 833--840, 2011.

\bibitem[\protect\citeauthoryear{Roweis and Saul}{2000}]{roweis2000nonlinear}
Sam~T Roweis and Lawrence~K Saul.
\newblock Nonlinear dimensionality reduction by locally linear embedding.
\newblock {\em Science}, 290(5500):2323--2326, 2000.

\bibitem[\protect\citeauthoryear{Scherzer and
  Weickert}{2000}]{scherzer2000relations}
Otmar Scherzer and Joachim Weickert.
\newblock Relations between regularization and diffusion filtering.
\newblock {\em Journal of Mathematical Imaging and Vision}, 12(1):43--63, 2000.

\bibitem[\protect\citeauthoryear{Tenenbaum \bgroup \em et al.\egroup
  }{2000}]{tenenbaum2000global}
Joshua~B Tenenbaum, Vin De~Silva, and John~C Langford.
\newblock A global geometric framework for nonlinear dimensionality reduction.
\newblock {\em Science}, 290(5500):2319--2323, 2000.

\bibitem[\protect\citeauthoryear{Vincent \bgroup \em et al.\egroup
  }{2010}]{VincentLLBM10}
Pascal Vincent, Hugo Larochelle, Isabelle Lajoie, Yoshua Bengio, and
  Pierre{-}Antoine Manzagol.
\newblock Stacked denoising autoencoders: Learning useful representations in a
  deep network with a local denoising criterion.
\newblock {\em Journal of Machine Learning Research}, 11:3371--3408, 2010.

\bibitem[\protect\citeauthoryear{Wan \bgroup \em et al.\egroup
  }{2013}]{wan2013regularization}
Li~Wan, Matthew Zeiler, Sixin Zhang, Yann~L Cun, and Rob Fergus.
\newblock Regularization of neural networks using dropconnect.
\newblock In {\em Proceedings of the 30th International Conference on Machine
  Learning (ICML-13)}, pages 1058--1066, 2013.

\bibitem[\protect\citeauthoryear{Wang and Tu}{2013}]{DBLP:conf/cvpr/WangT13}
Bo~Wang and Zhuowen Tu.
\newblock Sparse subspace denoising for image manifolds.
\newblock In {\em 2013 {IEEE} Conference on Computer Vision and Pattern
  Recognition, Portland, OR, USA, June 23-28, 2013}, pages 468--475, 2013.

\bibitem[\protect\citeauthoryear{Yan \bgroup \em et al.\egroup
  }{2007}]{yan2007graph}
Shuicheng Yan, Dong Xu, Benyu Zhang, Hong-Jiang Zhang, Qiang Yang, and Stephen
  Lin.
\newblock Graph embedding and extensions: a general framework for
  dimensionality reduction.
\newblock {\em Pattern Analysis and Machine Intelligence, IEEE Transactions
  on}, 29(1):40--51, 2007.

\end{thebibliography}

\end{document}